\documentclass[conference]{IEEEtran}
\IEEEoverridecommandlockouts
\usepackage{cite}
\usepackage{amsmath,amssymb,amsfonts}
\usepackage{subcaption}

\usepackage{algorithm2e}
\RestyleAlgo{ruled}
\SetKwComment{Comment}{/* }{ */}
\SetKwInOut{Input}{Input}

\usepackage{graphicx}
\usepackage{multirow}
\usepackage{textcomp}
\usepackage{algpseudocode}
\usepackage{booktabs}
\usepackage{xcolor}
\usepackage{pifont}
\usepackage{hyperref}
\def\BibTeX{{\rm B\kern-.05em{\sc i\kern-.025em b}\kern-.08em
    T\kern-.1667em\lower.7ex\hbox{E}\kern-.125emX}}
\begin{document}

\title{Mapping Urban Population Growth from Sentinel-2 MSI and Census Data Using Deep Learning: A Case Study in Kigali, Rwanda\\
\thanks{This work was supported in part by the Swedish National Space Agency under Grant dnr 155/15, the Digital Futures under the grant for the EO-AI4GlobalChange Project, and in part by ESA-China Dragon 5 Program under the EO-AI4Urban Project.}
}

\author{\IEEEauthorblockN{Sebastian Hafner}
\IEEEauthorblockA{\textit{KTH Royal Inst. of Technology}\\
Stockholm, Sweden \\
shafner@kth.se}
\and
\IEEEauthorblockN{Stefanos Georganos}
\IEEEauthorblockA{\textit{KTH Royal Inst. of Technology}\\
Stockholm, Sweden \\
stegeo@kth.se}
\and
\IEEEauthorblockN{Theodomir Mugiraneza}
\IEEEauthorblockA{\textit{University of Rwanda}\\
Kigali, Rwanda \\
thmugiraneza@gmail.com}
\and
\IEEEauthorblockN{Yifang Ban}
\IEEEauthorblockA{\textit{KTH Royal Inst. of Technology}\\
Stockholm, Sweden \\
yifang@kth.se}
}

\maketitle

\begin{abstract}

To better understand current trends of urban population growth in Sub-Saharan Africa, high-quality spatiotemporal population estimates are necessary. While the joint use of remote sensing and deep learning has achieved promising results for population distribution estimation, most of the current work focuses on fine-scale spatial predictions derived from single date census, thereby neglecting temporal analyses. In this work, we focus on evaluating how deep learning change detection techniques can unravel temporal population dynamics at short intervals. Since Post-Classification Comparison (PCC) methods for change detection are known to propagate the error of the individual maps, we propose an end-to-end population growth mapping method. Specifically, a ResNet encoder, pretrained on a population mapping task with Sentinel-2 MSI data, was incorporated into a Siamese network. The Siamese network was trained at the census level to accurately predict population change. The effectiveness of the proposed method is demonstrated in Kigali, Rwanda, for the time period 2016--2020, using bi-temporal Sentinel-2 data. Compared to PCC, the Siamese network greatly reduced errors in population change predictions at the census level. These results show promise for future remote sensing-based population growth mapping endeavors. Code is available on GitHub\footnote{\url{https://github.com/SebastianHafner/PopulationGrowthMapping_Kigali.git}}.
\end{abstract}


\section{Introduction}

The projections in the World Population Prospects 2022 report suggest that the global population could reach 9.7 billion in 2050 \cite{un2022world}. At the forefront of the anticipated population growth are countries of Sub-Saharan Africa. In light of this, frequent updates of existing population data in that region are crucial, particularly considering that knowledge of population distribution is a necessary requisite for a wide range of applications. For example, population distribution maps provide vital information for vaccination campaigns, disaster response deployment, and urban mobility and transport planning.

In recent years, census-independent (i.e., bottom-up) population mapping using deep learning and satellite imagery has shown promise in providing accurate population estimates. For example, Doupe \textit{et al.} \cite{doupe2016equitable} mapped population density at 8 km spatial resolution in Tanzania and Kenya using a Convolutional Neural Network (CNN) based on the VGG architecture and Landsat 7 imagery. Landsat 7 imagery and the VGG-net were also used by Robinson \textit{et al.} \cite{robinson2017deep} to predict population counts in the United States at 1 km spatial resolution. Authors in \cite{hu2019mapping} proposed to fuse Landsat 8 optical data with Sentinel-1 radar data to predict population density at 4.5 km spatial resolution for rural villages in India and demonstrated that dual-branch fusion networks outperform uni-modal networks. Sentinel-2 (S2) MultiSpectral Instrument (MSI) imagery was used by Huang \textit{et al.} \cite{huang2021sensing} to map population distribution at 1 km spatial resolution for the Atlanta, Georgia, and Dallas, Texas metropolitan areas in the United States of America. Recently, Neal \textit{et al.} \cite{neal2022census} used WorldView-2 imagery for estimating population in two districts of  Mozambique using representation learning. A ResNet was also used in \cite{georganos2022census} to map population in Sub-Saharan African cities with multisource satellite imagery from Pleiades and S2. Building footprints were further used to improve the geographical transferability of models.

While deep learning-based population mapping from satellite imagery has gained traction in recent years \cite{neal2022census, georganos2022census, huang2021sensing, hu2019mapping, robinson2017deep, doupe2016equitable}, little attention has been paid to population growth mapping with the exception of \cite{zhuang2021mapping}. Using a ResNet and Landsat 5 imagery, Zhuang \textit{et al.} \cite{zhuang2021mapping} performed population growth analysis in China for the 1985-2010 period by mapping population distribution at 1 km spatial resolution with a 5-year interval. However, analyzing population growth by comparative analysis of independently produced population maps, i.e., change detection by Post-Classification Comparison (PCC), is well-known to suffer from the error propagation of the individual population maps. To that end, we propose an end-to-end population growth mapping method to overcome the error propagation of PCC in uni-temporal population maps. This study is, up to the best of our knowledge, the first to map population growth in an end-to-end fashion from satellite imagery.

\section{Study Area and Data}
\label{sec:studyarea_data}

Kigali, the capital city and economic hub of Rwanda, was selected as the study area. Kigali encompasses an area of approximately 730 km$^2$. In 2012, Kigali had a population of approximately 1.1 million and placed among the fastest-growing cities in Africa \cite{nisr2012population}. In recent years, rapid urbanization resulted in the conversion of major cropland areas into built-up areas in the urban fringe zones of Kigali, which increased ecosystem service demands and negatively affected the habitat for biodiversity service function \cite{mugiraneza2022monitoring}.

\begin{figure}
     \centering
     \begin{subfigure}[b]{0.24\textwidth}
         \centering
         \includegraphics[width=\textwidth]{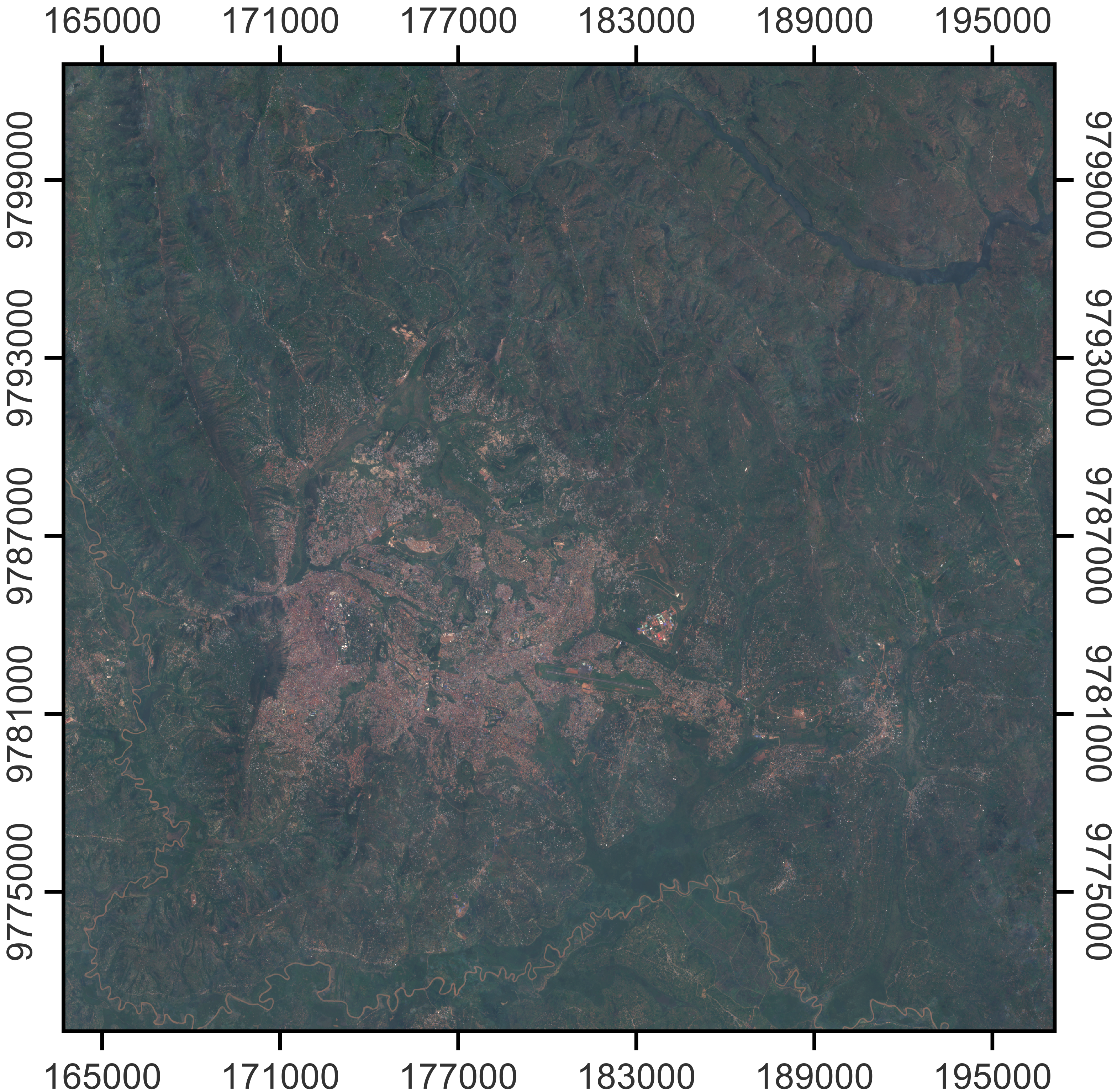}
         \caption{}
         \label{subfig:s2_2016_wet}
     \end{subfigure}
     \hfill
     \begin{subfigure}[b]{0.24\textwidth}
         \centering
         \includegraphics[width=\textwidth]{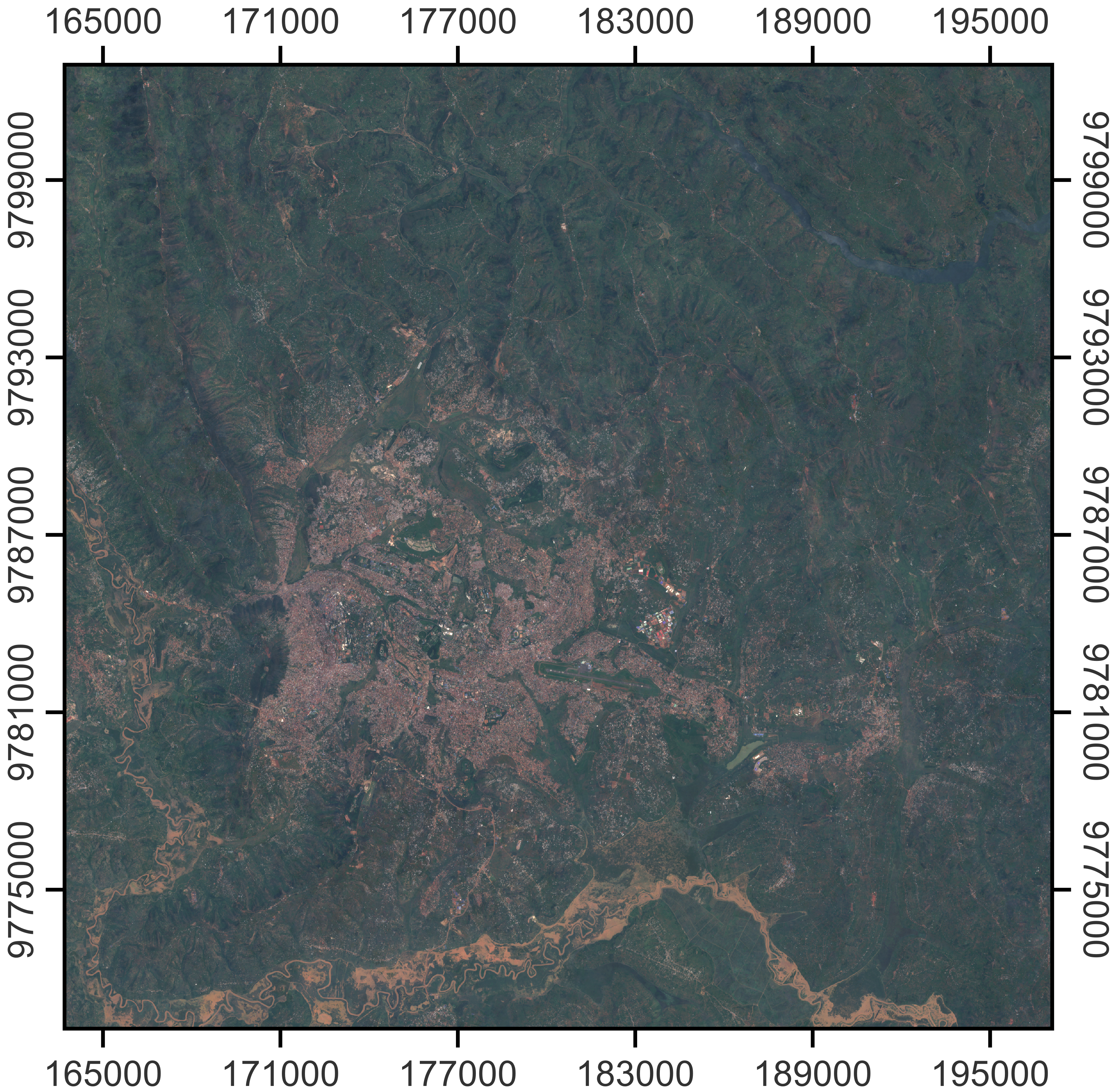}
         \caption{}
         \label{subfig:s2_2020_wet}
     \end{subfigure}
     \hfill
     \begin{subfigure}[b]{0.24\textwidth}
         \centering
         \includegraphics[width=\textwidth]{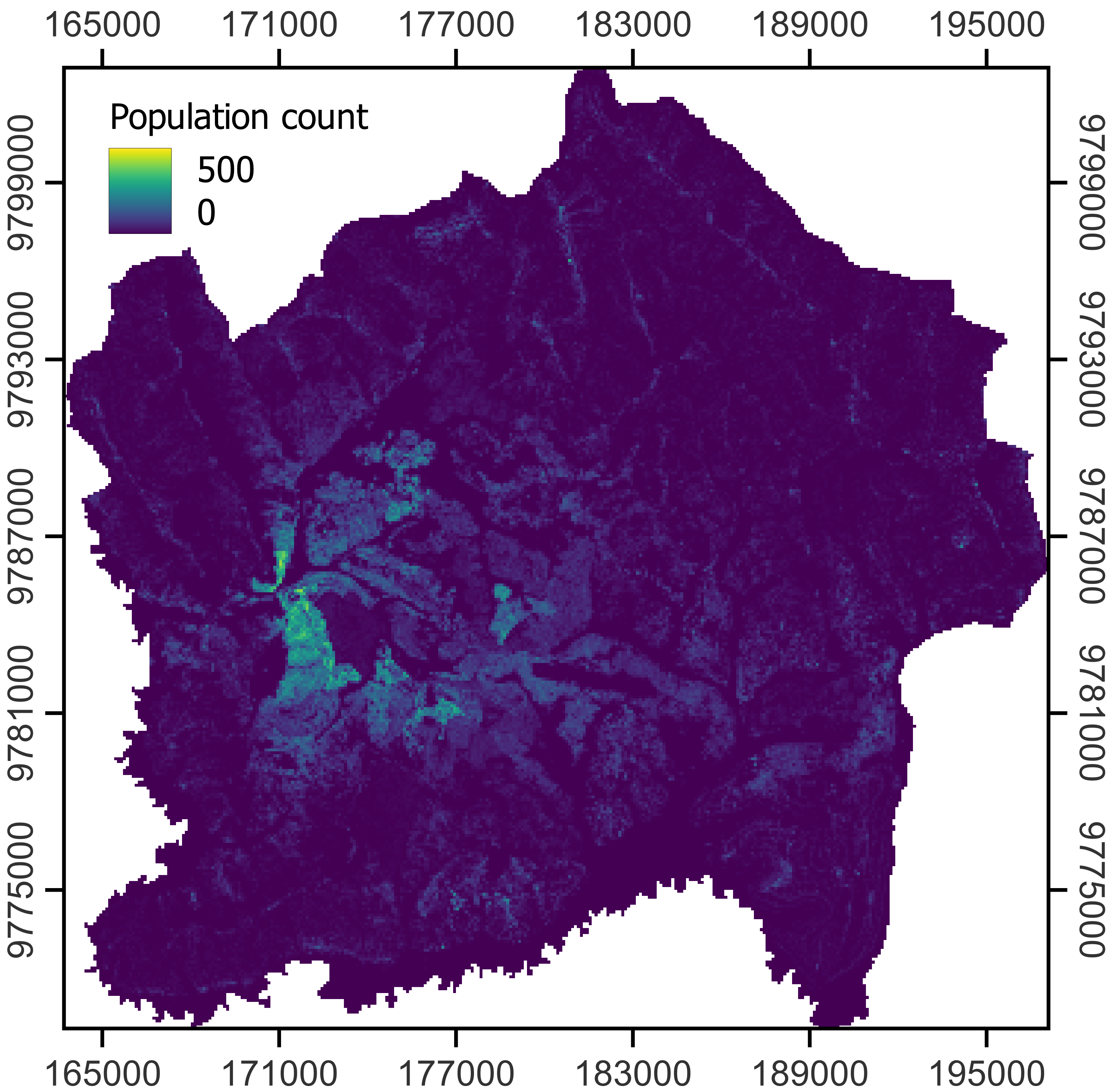}
         \caption{}
         \label{subfig:pop_label_2020}
     \end{subfigure}
     \hfill
     \begin{subfigure}[b]{0.24\textwidth}
         \centering
         \includegraphics[width=\textwidth]{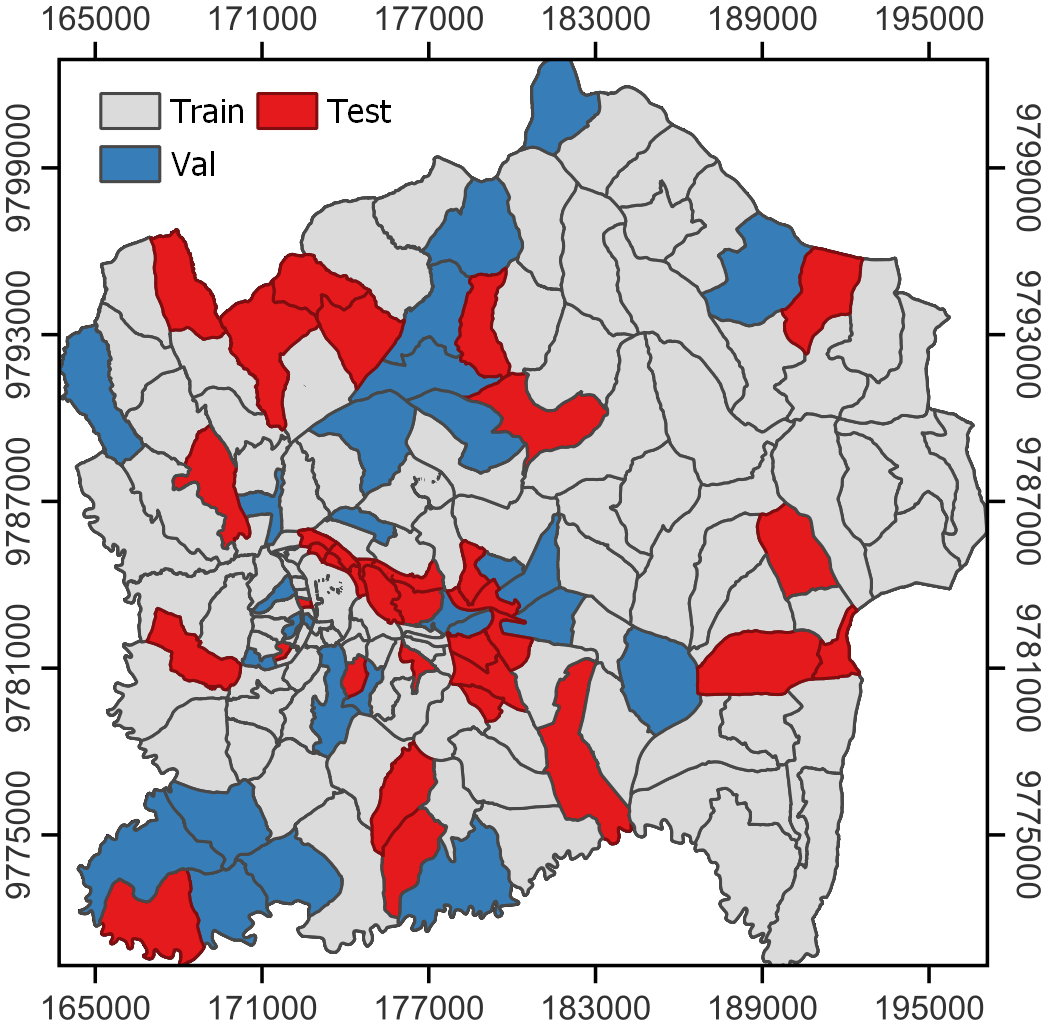}
         \caption{}
         \label{subfig:train_test_split}
     \end{subfigure}
     \hfill
        \caption{S2 MSI composites for (a) 2016 and (b) 2020, and (c) population labels at grid level. (d) shows the data set splits.}
        \label{fig:study_area}
\end{figure}

S2 MSI imagery of Kigali for 2016 and 2020 was retrieved from Google Earth Engine \cite{gorelick2017google}. Specifically, cloud-free composites were generated by collecting all S2 Level-1C (top-of-atmosphere) scenes acquired during the wet season of the respective year. Thereafter, cloudy pixels (i.e., cloud probability $>$ 50 \%) were masked for each scene, before the scenes were combined using median compositing. The resulting cloud-free composites for 2016 and 2020 are visualized in Figure \ref{subfig:s2_2016_wet} and Figure \ref{subfig:s2_2020_wet}, respectively.

Population census data at the level of designated census enumeration areas were acquired for Kigali for the years 2016 and 2020 (161 administrative polygons). These areas are corresponding to the smallest administrative entities in Rwanda called villages. The data consist of number of population (head counts) and were acquired from  two institutions including Kigali city One Stop center and the Local Administrative Entities Development Agency. Using iterative merging, we aggregated the dataset into a smaller number of units, to reflect a more realistic scenario regarding data availability, but also to adapt to the needs of the experiment (i.e., 100 meter predictive spatial resolution).
Finally, the census units were randomly split into a training, validation, and test set (60/20/20 split) (Figure \ref{subfig:train_test_split}).

\section{Methodology}

\subsection{Problem Setup}

We consider two S2 MSI images that cover the same geographical area (Kigali) but were acquired at two different times, $t_1$ and $t_2$. Furthermore, we consider the census units constituting the City of Kigali, where each census unit, $U$, contains an accurate count of the population, $Y$, for $t_1$ and $t_2$. The goal is to train a network that accurately predicts the population growth for a census unit $D$ ($= Y^{\rm t_2} - Y^{\rm t_1}$) from the part of the S2 images $I^{\rm t_1}$ and $I^{\rm t_2}$ covering $U$. However, each census unit has a unique non-rectangular shape and, therefore, cannot be used directly as network input. A common way to deal with this is to operate on a grid level by dividing the entire study area into patches that constitute the census areas \cite{georganos2022census}. Consequently, census units are composed of a varying number of patches (100 x 100 m). Therefore, the network input to predict the population growth for a census unit is, in practice, the collection of S2 patches, $x^{t1}$ and $x^{t2}$, constituting the census unit.

\subsection{Proposed Method}

The proposed population growth mapping method consists of two stages: 1) an encoder model is pretrained by mapping population at the grid level, and 2) a Siamese network, incorporating the pretrained encoder, is trained at the census level to map population growth. 

\paragraph*{Population Mapping at Grid Level}
\label{subsec:pop_mapping}

\begin{figure}[h]
    \centering
    \includegraphics[width=.48\textwidth]{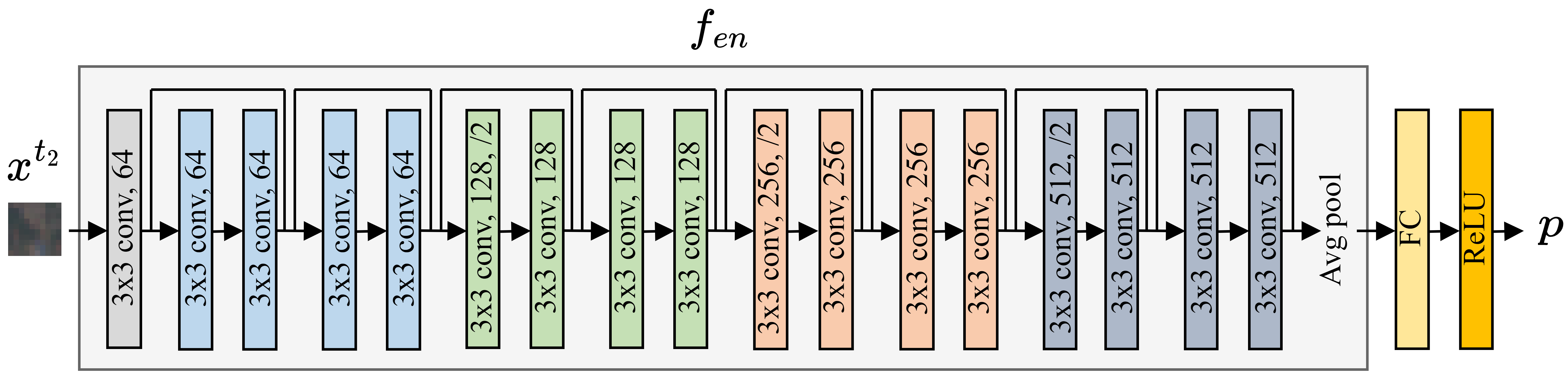}
    \caption{Diagram of the ResNet-18 model used for grid-level population mapping.}
    \label{fig:pretraining}
\end{figure}

Our previous work demonstrated that an encoder based on the ResNet-18 architecture suffices to learn salient features from S2 MSI imagery for population mapping \cite{georganos2022census}. The same architecture is employed in this work (Figure \ref{fig:pretraining}). Specifically, the first layer of the ResNet-18 encoder is replaced with a 3 x 3 conv layer with 4 input channels to accommodate the 10 m S2 bands (Band 2, 3, 4, and 8) as input, while the remaining conv blocks constituting the encoder remain unchanged. The features extracted with the encoder are converted to a population prediction, $p$, using a fully connected layer. Finally, the ReLu activation function is used to constrain $p$ values to positive numbers.

Hyper-parameters for training are tuned on the validation set using grid search with 3 learning rates ($10^{-5}$, $10^{-4}$, and $10^{-3}$) and 2 batch sizes (8, 16). AdamW is used as optimizer, and the training duration is set to 100 epochs with early stopping (patience 5) to prevent models from overfitting to the training set. As in \cite{georganos2022census}, flips (horizontal and vertical) and rotations ($k * 90^{\circ}$, where $k \in \{0, 1, 2, 3\}$) are applied to the training data for data augmentation, and the Mean Square Error (MSE) loss (commonly known as L2 loss) is used as loss function. L2 loss is defined as follows: $L2 = (y - p)^2$, where the true and predicted population value is denoted by $y$ and $p$, respectively. An NVIDIA GeForce RTX 3090 graphics card is used for training.

\paragraph*{Population Growth Mapping at Census Level}
\label{subsec:pop_growth_mapping}

\begin{figure}[h]
    \centering
    \includegraphics[width=.48\textwidth]{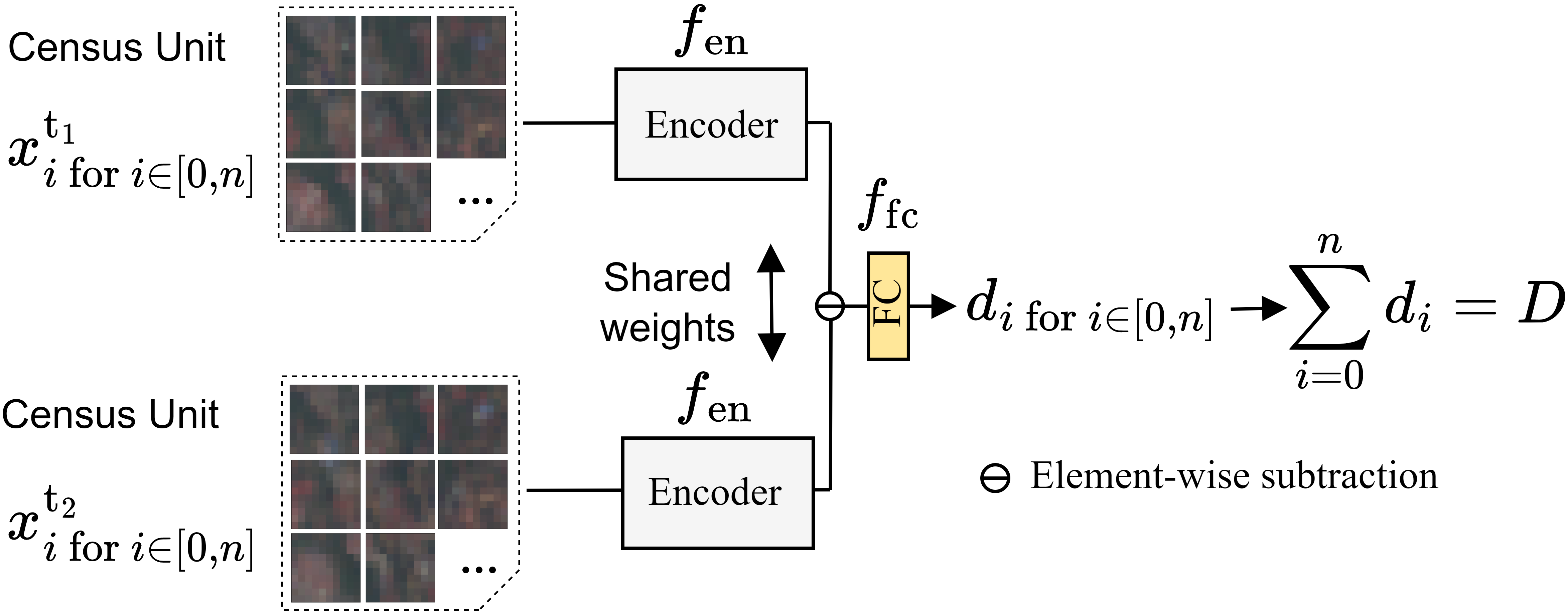}
    \caption{Diagram of the proposed population growth mapping method consisting of two pretrained ResNet-18 encoders, $f_{\rm en}$, with shared weights and a fully connected layer, $f_{\rm fc}$. The network is trained at the census level with frozen encoders.}
    \label{fig:proposed_method}
\end{figure}

For population growth mapping, we incorporate the pretrained ResNet-18 encoder into a Siamese network (Figure \ref{fig:proposed_method}). Siamese networks consist of two encoders with shared weights that are used to separately extract features from the inputs, before deriving the change information from the combined features. Due to their inherent suitability to detect differences, Siamese networks have also become a popular architecture for change detection in bi-temporal pairs of satellite images. In this work, the pretrained encoder is employed to extract features on population count from both images separately. The pair of bi-temporal features is then converted to a population growth prediction using a fully connected layer. No activation function is applied to the output of that layer to allow for negative growth predictions.

An important challenge of supervised population growth mapping is that bi-temporal population counts are required for the derivation of growth labels. While it is possible to accurately disaggregate a census to a grid, this requires auxiliary data such as land cover maps or building footprints. However, this data is often not available for both timestamps. Therefore, the Siamese network is trained at the census level by adapting the weakly supervised learning strategy proposed in \cite{metzger2022fine}. Specifically, Metzger \textit{et al.} \cite{metzger2022fine} trained a population mapping model using population count at the census level as labels by comparing them to the aggregated model predictions (patch-level) for corresponding census units. Likewise, we use the Siamese network to predict population growth separately for all patches of a census unit, before applying the loss to the sum of predicted growth, $D$, using $\Delta Y$ as label. The training setup (i.e., hyper-parameter tuning, early stopping, and data augmentations) is identical to that for population mapping. It should be noted, however, that the pretrained encoder is frozen during training, meaning that only the fully connected layer ($f_{\rm fc}$ in Figure \ref{fig:proposed_method}) is trained.

\subsection{Accuracy Metrics}

We make use of three commonly employed metrics in population studies \cite{linard2012population}, namely the Root Mean Squared Error (RMSE), the Mean Absolute Error (MAE), and the coefficient of determination (R$^2$). RMSE and MAE are defined as follows:

\begin{equation}
\label{eq:RMSE}
\text{RMSE} = \sqrt{\frac{\sum^n_{i=1}(y_i - p_i)^2}{n}}, \ \text{MAE} = \sqrt{\frac{\sum^n_{i=1}\left | y_{i} - p_{i} \right |}{n}},
\end{equation}

where $y$ and $p$ are true and predicted values, respectively, and $n$ is the sample size. On the other hand, R$^2$ is defined as 1 minus the fraction of the  residual sum of squares and the total variability of the data.

\section{Results}

Table \ref{tab:quan_popmapping_results} lists the quantitative population mapping results at the grid level for 2020 and at the census level for 2016 and 2020. All three accuracy metrics indicate that accurate population predictions were achieved at the grid level. However, the aggregated results at the census level provide a stronger validation since the census population counts are official data. While RMSE and MAE values are not comparable between the grid and census level, the R$^2$ values at the census level indicate good performance (0.70 +), although worse than the performance achieved at the grid level (0.84). It is also apparent that the obtained accuracy values for 2016 and 2020 are relatively similar. Consequently, applying the model to new data from a different year had little impact on model performance.

\begin{table}[h]
\small
  \caption{Quantitative population mapping results at the gird and census level for the test set.}
  \label{tab:quan_popmapping_results}
  \centering
  \begin{tabular}{lrrrrrrrrr}
    \toprule
        \multirow{2}{3em}{Level} & \multicolumn{2}{c}{RMSE $\downarrow$} && \multicolumn{2}{c}{MAE $\downarrow$} && \multicolumn{2}{c}{$\text{R}^2$ $\uparrow$} \\
    & 2016 & 2020 && 2016 & 2020 && 2016 & 2020 \\
    \cmidrule{2-3} \cmidrule{5-6} \cmidrule{8-9}
    Grid & - & 19 && - & 10 && - & 0.84  \\
    Census & 3,199 & 3,253 && 2,368 & 2,196 && 0.72 & 0.73  \\
    \bottomrule
  \end{tabular}
\end{table}

\begin{figure}
     \centering
     \begin{subfigure}[b]{0.24\textwidth}
         \centering
         \includegraphics[width=\textwidth]{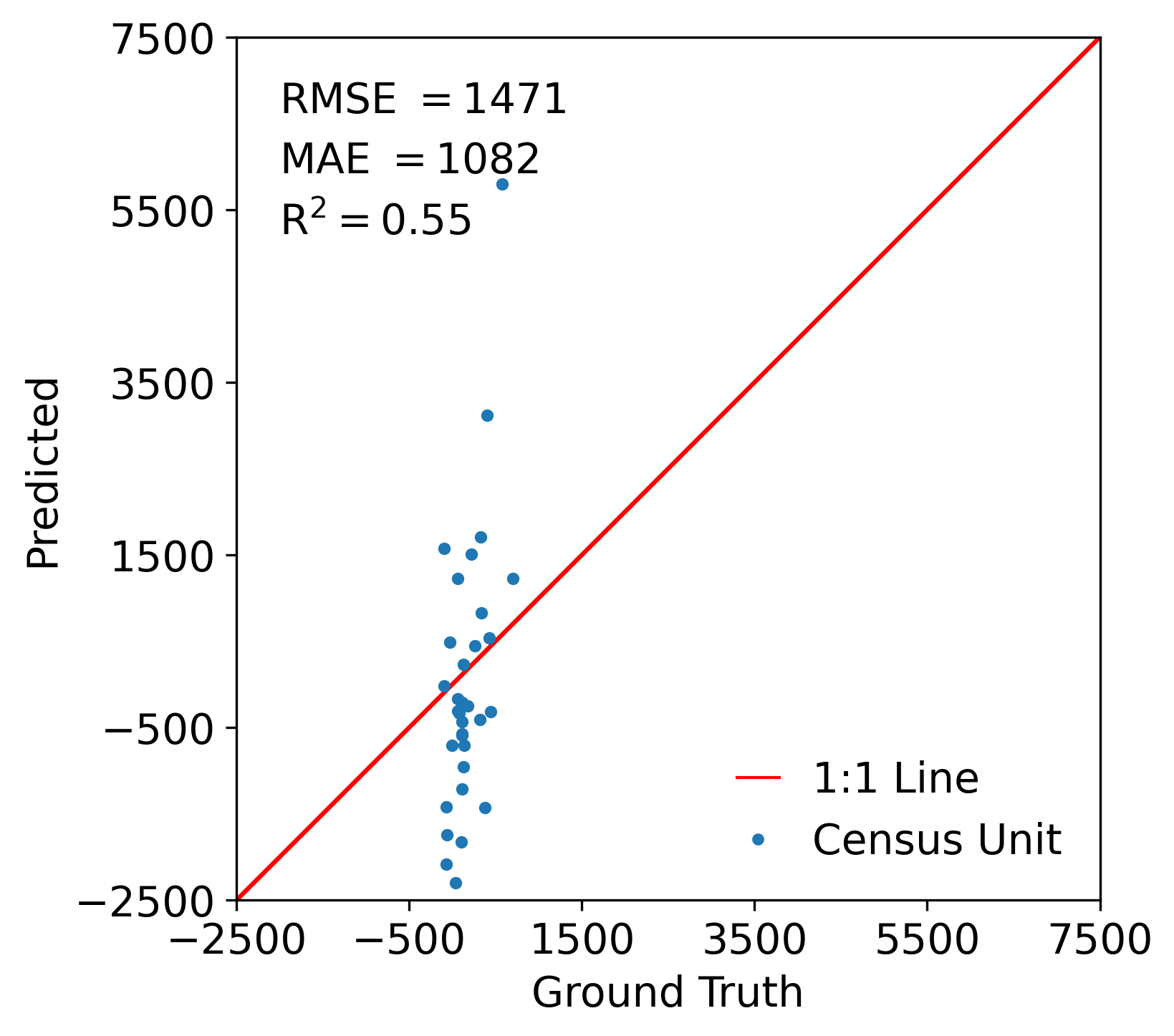}
         \caption{}
         \label{subfig:quan_pcc}
     \end{subfigure}
     \hfill
     \begin{subfigure}[b]{0.235\textwidth}
         \centering
         \includegraphics[width=\textwidth]{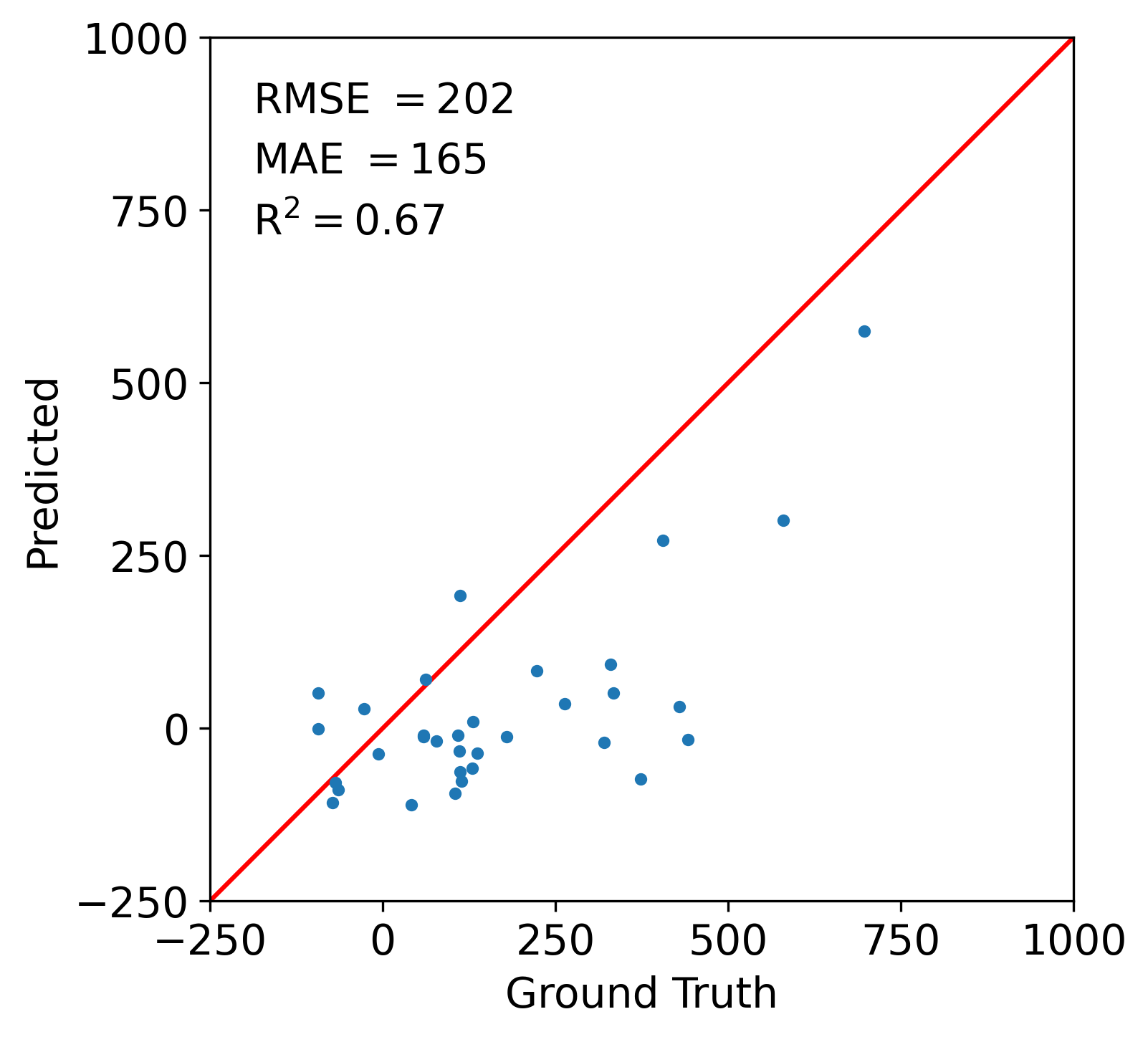}
         \caption{}
         \label{subfig:quan_ours}
     \end{subfigure}
        \caption{Population growth test results at the census level for (a) PCC and (b) the proposed method.}
        \label{fig:quan_growthmapping}
\end{figure}

Figure \ref{fig:quan_growthmapping} quantitatively compares the population growth predictions of (a) the PCC with (b) the proposed end-to-end method. The former, PCC, performed poorly, resulting in very high errors (RMSE = 1,471 and MAE = 1,082). In contrast, the proposed method achieved satisfactory results with an RMSE of 202 and an MAE of 165. In terms of $\text{R}^2$ values, the results are more similar, but better performance was also achieved by the proposed method (0.55 vs. 0.67). However, it is also apparent that the proposed method generally underestimates population growth.

\begin{figure}
     \centering
     \begin{subfigure}[b]{0.48\textwidth}
         \centering
         \includegraphics[width=\textwidth]{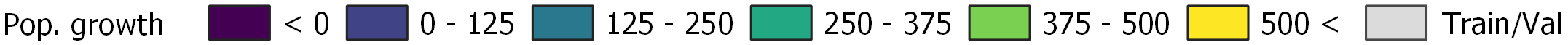}
         \label{subfig:label}
     \end{subfigure}
     \begin{subfigure}[b]{0.24\textwidth}
         \centering
         \includegraphics[width=\textwidth]{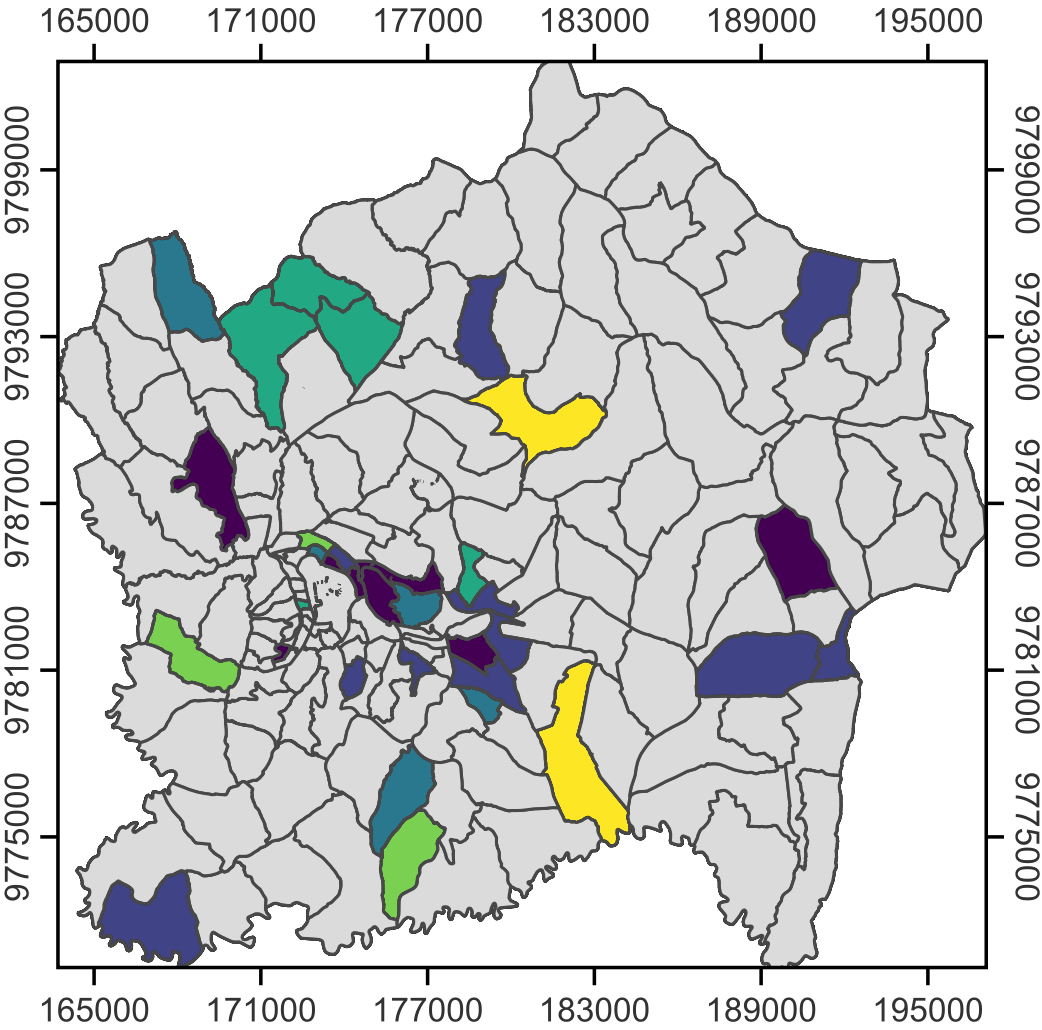}
         \caption{}
         \label{subfig:qual_gt}
     \end{subfigure}
     \begin{subfigure}[b]{0.24\textwidth}
         \centering
         \includegraphics[width=\textwidth]{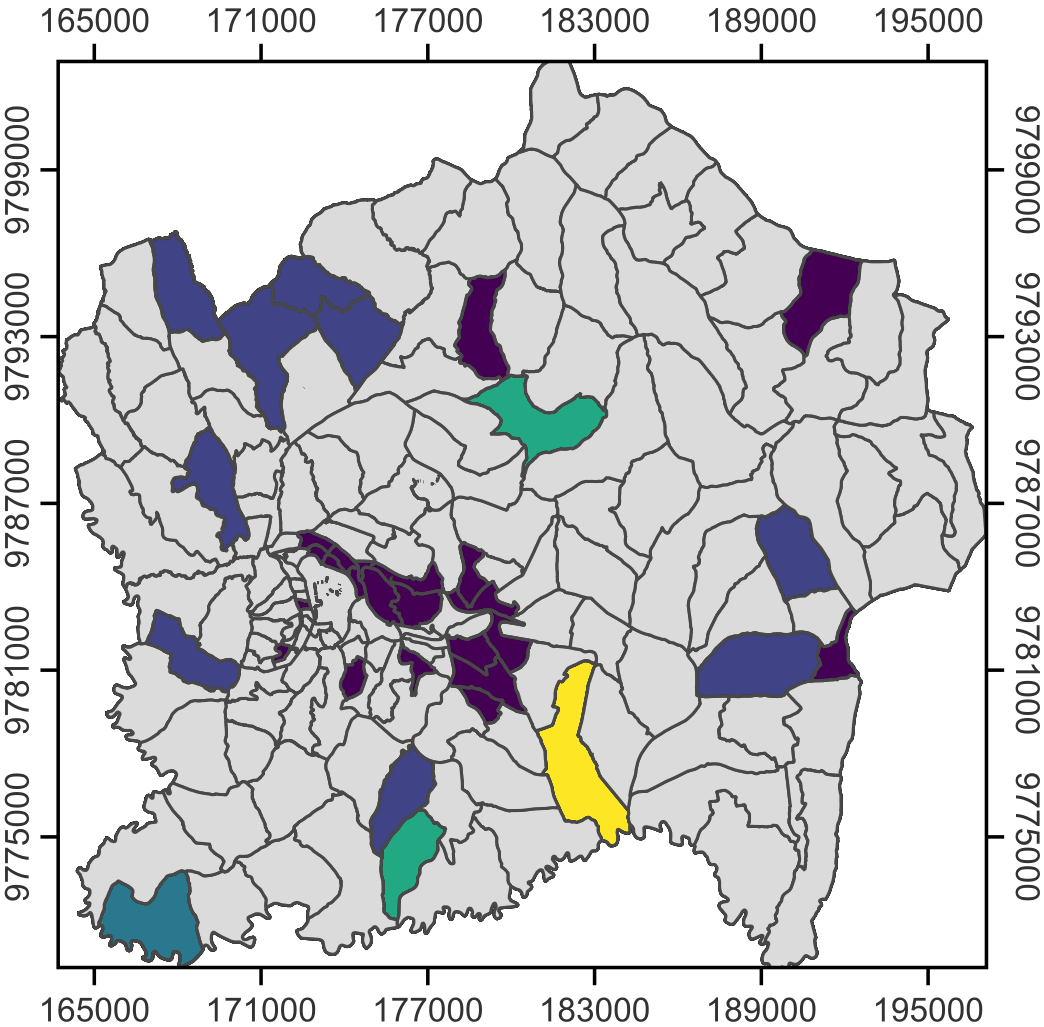}
         \caption{}
         \label{subfig:qual_ours}
     \end{subfigure}
        \caption{Population growth maps for the test census units. (a) shows the ground truth and (b) our predictions.}
        \label{fig:qual_test_results}
\end{figure}

The qualitative population growth mapping predictions of the proposed method are visualized in Figure \ref{subfig:qual_ours}, next to the ground truth in Figure \ref{subfig:qual_gt}. Although the magnitude of growth was underestimated, the proposed method picked up on the population growth that occurred on the outskirts of Kigali (e.g., in the northeast and in the central south). However, the model failed to detect population growth in the small census units of central Kigali for which it predicted slightly negative growth values.

\section{Discussion and Limitations}

We find the proposed method to be effective for population growth mapping from S2 MSI imagery, especially compared to PCC. Our findings also emphasize that salient features about population count can be learned from S2 imagery using a ResNet model. These results are in line with \cite{georganos2022census}.

Our work is also subject to several limitations. First of all, to train the Siamese network, bi-temporal census data is required. However, census data, let alone bi-temporal census data, is difficult to obtain in Sub-Saharan Africa, or often not available at all \cite{linard2012population}. Moreover, the S2 mission was launched less than 8 years ago, while censuses are typically conducted every 10 years. Consequently, bi-temporal census data for time periods starting after 2015 are largely unavailable. Another limitation of this work is that population predictions are based on the presence of built-up areas, but the land use of these areas may not be residential \cite{robinson2017deep}. To overcome this, Neal \textit{et al.} \cite{neal2022census} suggest including additional data modalities like, for example, night-time light data. Our quantitative results in central Kigali (Figure \ref{subfig:qual_ours}) also suggest that densification of urban areas, and the subsequent increase in population, may be challenging to accurately predict. Finally, further work is needed to assess if the proposed method can accurately detect negative population growth as a result of, for example, slum evictions.

\section{Conclusion}

In this paper, a population growth mapping method based on a Siamese network is proposed and evaluated in Kigali, Rwanda for the time period 2016--2020. Using S2 MSI data as input, the proposed method achieved satisfactory population growth mapping results at the census level (RMSE = 202, MAE = 165, $\text{R}^2$ = 0.67), and greatly outperformed PCC in terms of RMSE (-1,269) and MAE (-917).

Our future work will extend the study area to other Sub-Saharan African cities. Furthermore, we will investigate semi-supervised learning for Siamese network training (e.g., \cite{hafner2022urban}) to reduce the dependence on bi-temporal census data.

\bibliography{ref.bib}

\begin{thebibliography}{10}

\bibitem{un2022world}
{United Nations Department of Economic and Social Affairs, Population
  Division},
\newblock ``World population prospects 2022: Summary of results,''
\newblock Tech. {R}ep. UN DESA/POP/2022/TR/NO. 3, 2022.

\bibitem{doupe2016equitable}
Doupe, P. et~al,
\newblock ``Equitable development through deep learning: The case of
  sub-national population density estimation,''
\newblock in {\em Proceedings of the 7th Annual Symposium on Computing for
  Development}, 2016, pp. 1--10.

\bibitem{robinson2017deep}
Robinson, C. et~al,
\newblock ``A deep learning approach for population estimation from satellite
  imagery,''
\newblock in {\em Proceedings of the 1st ACM SIGSPATIAL Workshop on Geospatial
  Humanities}, 2017, pp. 47--54.

\bibitem{hu2019mapping}
Hu, W. et~al,
\newblock ``Mapping missing population in rural india: A deep learning approach
  with satellite imagery,''
\newblock in {\em Proceedings of the 2019 AAAI/ACM Conference on AI, Ethics,
  and Society}, 2019, pp. 353--359.

\bibitem{huang2021sensing}
Huang, X. et~al,
\newblock ``Sensing population distribution from satellite imagery via deep
  learning: Model selection, neighboring effects, and systematic biases,''
\newblock {\em IEEE Journal of Selected Topics in Applied Earth Observations
  and Remote Sensing}, vol. 14, pp. 5137--5151, 2021.

\bibitem{neal2022census}
Neal, I. et~al,
\newblock ``Census-independent population estimation using representation
  learning,''
\newblock {\em Scientific Reports}, vol. 12, no. 1, pp. 1--12, 2022.

\bibitem{georganos2022census}
Georganos, S. et~al,
\newblock ``A census from heaven: Unraveling the potential of deep learning and
  earth observation for intra-urban population mapping in data scarce
  environments,''
\newblock {\em International Journal of Applied Earth Observation and
  Geoinformation}, vol. 114, pp. 103013, 2022.

\bibitem{zhuang2021mapping}
Zhuang, H. et~al,
\newblock ``Mapping multi-temporal population distribution in china from 1985
  to 2010 using landsat images via deep learning,''
\newblock {\em Remote Sensing}, vol. 13, no. 17, pp. 3533, 2021.

\bibitem{nisr2012population}
{National Institute of Statistics of Rwanda},
\newblock ``The 2012 population and housing census results,''
\newblock Tech. {R}ep., 2012.

\bibitem{mugiraneza2022monitoring}
Mugiraneza, T. et~al,
\newblock ``Monitoring urbanization and environmental impact in kigali, rwanda
  using sentinel-2 msi data and ecosystem service bundles,''
\newblock {\em International Journal of Applied Earth Observation and
  Geoinformation}, vol. 109, pp. 102775, 2022.

\bibitem{gorelick2017google}
Gorelick, N. et~al,
\newblock ``Google earth engine: Planetary-scale geospatial analysis for
  everyone,''
\newblock {\em Remote sensing of Environment}, vol. 202, pp. 18--27, 2017.

\bibitem{metzger2022fine}
Metzger, N. et~al,
\newblock ``Fine-grained population mapping from coarse census counts and open
  geodata,''
\newblock {\em Scientific Reports}, vol. 12, no. 1, pp. 20085, Nov 2022.

\bibitem{linard2012population}
Linard, C. et~al,
\newblock ``Population distribution, settlement patterns and accessibility
  across africa in 2010,''
\newblock {\em PloS one}, vol. 7, no. 2, pp. e31743, 2012.

\bibitem{hafner2022urban}
Hafner, S. et~al,
\newblock ``Urban change detection using a dual-task siamese network and
  semi-supervised learning,''
\newblock in {\em IGARSS 2022-2022 IEEE International Geoscience and Remote
  Sensing Symposium}. IEEE, 2022, pp. 1071--1074.

\end{thebibliography}
\bibliographystyle{IEEEbib}

\end{document}